\documentclass[letterpaper]{article} 
\usepackage[preprint]{aaai24}  
\usepackage{times}  
\usepackage{helvet}  
\usepackage{courier}  
\usepackage[hyphens]{url}  
\usepackage{graphicx} 
\usepackage{natbib}  
\usepackage{caption} 
\usepackage{amsmath, amssymb}
\usepackage{algorithm}
\usepackage{algorithmic}
\usepackage{amsfonts}
\usepackage{xcolor}

\newcommand{\cl}{\mathcal}
\newcommand{\bb}{\mathbb}

\newcommand{\bo}{\boldsymbol}

\title{AutoIRT: Calibrating Item Response Theory Models with Automated Machine Learning}

\author{
    James Sharpnack\textsuperscript{\rm 1},
    Phoebe Mulcaire\textsuperscript{\rm 1},
    Klinton Bicknell\textsuperscript{\rm 1},
    Geoff LaFlair\textsuperscript{\rm 1},
    Kevin Yancey\textsuperscript{\rm 1},
}
\affiliations{
    \textsuperscript{\rm 1}Duolingo, Pittsburgh, PA 15206 \\
    \{james.sharpnack, phoebe, klinton, geoff, kyancey\}@duolingo.com
}

\begin{document}

\maketitle

\begin{abstract}
Item response theory (IRT) is a class of interpretable factor models that are widely used in computerized adaptive tests 
(CATs), such as language proficiency tests.
Traditionally, these are fit using parametric mixed effects models on the probability of a test taker getting the correct answer to a test item (i.e., question).
Neural net extensions of these models, such as BertIRT, require specialized architectures and parameter tuning.
We propose a multistage fitting procedure that is compatible with out-of-the-box Automated Machine Learning (AutoML) tools.
It is based on a Monte Carlo EM (MCEM) outer loop with a two stage inner loop, which trains a non-parametric AutoML grade model using item features followed by an item specific parametric model.
This greatly accelerates the modeling workflow for scoring tests.
We demonstrate its effectiveness by applying it to the Duolingo English Test, a high stakes, online English proficiency test.
We show that the resulting model is typically more well calibrated, gets better predictive performance, and more accurate scores than existing methods (non-explanatory IRT models and explanatory IRT models like BERT-IRT).
Along the way, we provide a brief survey of machine learning methods for calibration of item parameters for CATs.
\end{abstract}

\section{Introduction}

The purpose of any test is to measure a construct, or abstract characteristic, such as listening comprehension or speech production, and summarize the test takers proficiency in the form of a score (or scores).
High-stakes computerized adaptive tests (CATs) are designed with several goals in mind, and we outline a few here.
First, test designers seek to meet high reliability, validity, and security standards \cite{AERA2014}.
Second, ideally each item (i.e., question or exercise) given to a test taker provides the most amount of information about the test taker's proficiency as possible.
This is accomplished by making a test adaptive, for example, by estimating the ability of a test taker based on previous responses and then administering subsequent items with corresponding difficultly.
Third, it is important that the reported scores accurately reflect the test taker proficiency in the constructs that the test purports to measure.
This is largely accomplished by accurately calibrating the test item bank.
Calibrating items refers to measuring their psychometric characteristics, such as difficulty and discrimination, that can then be used for adaptive administration and scoring of the items.

Item response theory (IRT) models an item's psychometric characteristics, known as item parameters, via parametric models for a test taker's grade on a given item \cite{lord1980, wright1979best}.
A key advantage of IRT models is their interpretability, which is important for curating an item bank, e.g., ensuring that the item bank contains a wide range of difficulties, and for constructing CAT administration rules.
However, traditional IRT calibration requires many test taker responses (e.g., hundreds) for each item to meet the standards of a high-stakes test.
This data is typically collected in an item piloting phase prior to the launch of the calibrated items on the test.
Piloting has negative consequences because during this process the test takers are answering items that are not used for scoring and are not tailored to the test taker proficiency.
If piloting is performed outside of the high-stakes test, such as in practice tests, then this can pose security risks for the test because the items are exposed to the public \cite{laflair2022digital, way1998protecting}.
Furthermore, test takers have less motivation on practice tests, which is known to impact the psychometric characteristics of the items \cite{cheng2014motivation}.
One approach to calibrating new items with few responses is to use feature-based machine learning extensions of traditional IRT models \cite{fischer1973lltm, mccarthy2021jump, yancey2024bert}.
The features in this case come from the content of the item itself, and are constructed using NLP features or LLM embeddings (such as BERT embeddings).

In this work, we introduce a new approach to calibrating item parameters called AutoIRT.
AutoIRT enables the use of automated machine learning (AutoML) to train an IRT model from test response data and item content.
AutoML refers to general purpose tools for training and evaluating ML models.
They are designed to circumvent laborious hyperparameter tuning, feature engineering, architecture search, and to automatically support multi-modal input.
Examples include AutoGluon \cite{erickson2020autogluon}, H2o \cite{ledell2020h2o}, and Auto-sklearn 2.0 \cite{feurer2022auto}.
For a survey and benchmark comparison see \cite{zoller2021benchmark, gijsbers2024amlb}.
The issue is that out-of-the-box AutoML will not produce interpretable psychometric models such as IRT models.

This work constitutes the first use of AutoML to fit IRT models.
The fact that it produces IRT models makes it compatible with the prevailing approach in psychometrics for scoring and administering tests.
We provide an introduction to machine learning approaches to item response theory and outline prior work.
We demonstrate some key properties of AutoIRT by applying it to simulated data.
We apply our method to Duolingo English Test data from two task types, yes/no vocabulary and vocab-in-context, and compare it to prior approaches.

\section{Machine Learning Approaches to IRT}
\label{sec:irt}

In this section, we will provide a brief survey of item response theory (IRT) and ML approaches to IRT.

\subsection{Item Response Theory and Test Scoring}

The main byproduct of a test is a score (or scores) that reflects the test takers ability.
In IRT, this ability is modeled via a single parameter $\theta \in \bb R$ (i.e., the unidimensionality assumption).
The score is an estimate of the ability parameter; we will use the posterior expectation of $\theta$ given an uninformed prior and the grades.
In order to make $\theta$ more meaningful, IRT assumes that the expected grade is monotonic in $\theta$ (i.e., the monotonicity assumption).
It is also assumed that a test takers responses are independent given $\theta$ (i.e., the local independence assumption).
The most commonly used IRT models are the parametric logistic models for binary responses.
We denote the binary grades as $G_{i,s} \in \{0, 1\}$ (0 if incorrect, 1 if correct) that session $s$ obtained on item $i$.
For the vast majority of items, this value is missing, and we denote the set of response pairs $(i,s)$ as $\cal R$.

\begin{figure}
    \centering
    \includegraphics[width=\linewidth]{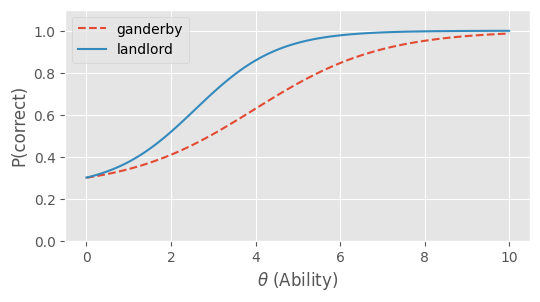}
    \caption{Example IRFs for two Y/N Vocab words fitted using AutoIRT, $\theta$ is on a $(0,10)$ scale.  The item "ganderby" has $a = 0.666, d = 3.969, c = 0.25$, and "landlord" has $a=1.021, d=2.572, c=0.25$.}
    \label{fig:example_irf}
\end{figure}

In these models the grade probability, also known as the item response function (IRF), takes the form,
\begin{equation}
\label{eq:3PL}
\begin{split}
    p(\theta, \phi_i) &= \bb P \{ G_{i,s} = 1 | \theta_s = \theta \} \\
    &= c_i + (1-c_i) \cdot \sigma (a_i (\theta - d_i)),    
\end{split}
\end{equation}
where $\phi_i = (a_i, c_i, d_i)$ are the item parameters: slope, or discrimination, $a_i$, chance parameter, $c_i$, and difficulty, $d_i$.
As the discrimination, $a_i$, increases, the IRF looks more like a step function with a change-point at $d_i$, meaning that it gives us more information about whether or not $\theta$ exceeds $d_i$.
This information content is diminished when the chance parameter $c_i$ increases.
In Figure \ref{fig:example_irf}, we see example IRFs for two Y/N Vocab words in the Duolingo English Test (DET) practice test.

Throughout we will assume that there are explanatory features for the items, $\bo x_i \in \bb R^d$.
These may be some combination of neural net embeddings such as BERT \cite{devlin2019bert} or CLIP \cite{radford2021learning}, hand-crafted linguistic features, or other salient aspects of the item.
The semi-parametric mixed effects model assumes that the item parameters take the following form based on the item features $\bo x_i$:
\begin{align}
    \log(a_i) &= f_a(\bo x_i) + \delta^a_i, \nonumber \\ 
    d_i &= f_d (\bo x_i) + \delta^d_i,
    \label{eq:mixed_effects}
\end{align}
where $f_a, f_d$ are functions mapping $\bb R^d \to \bb R$.
The terms $\delta^a_i, \delta^d_i \in \bb R$ are called the random effects terms, and typically can be represented using a one-hot encoding of the item id. 
If we ignore features completely ($f=0$) and only model the item parameters using random effects terms, then we call this a \emph{non-explanatory IRT model}.

In general, model \eqref{eq:3PL} can either have a free chance parameter, $c_i$, that needs to be fit, or a fixed chance parameter, $c_i = c$, that is assumed to be known.
When $c = 0$ is assumed, \eqref{eq:3PL} is the 2-parameter logistic (2PL) model, and otherwise it is the 3-parameter logistic (3PL) model.
There is also the commonly used 1PL model, which is also known as the Rasch model, where there is only the difficulty parameter and $a_i = 1, c_i = 0$.
Fixing the chance parameter is common because allowing $\theta$ and $c_i$ to simultaneously be free parameters makes the model unidentifiable \cite{baker2004item}.

Non-explanatory models can be calibrated in batch with maximum likelihood \cite{baker2004item}, marginal maximum likelihood \cite{bock1981mml}, or fully Bayesian MCMCs \cite{patz1999applications}.
The Elo rating system \cite{elo1978rating}, widely used in the chess world, is an online calibration procedure for the Rasch model \cite{baker1992item, kiraly2017elo}.
Calibrating item parameters for CATs is primarily done in batch after a sufficient number of test taker responses have been collected in a piloting phase.
After the piloting and calibration phases, those items can be used for scoring tests; during the pilot those items are administered to the test taker but not scored.
The score is a final estimate of $\theta_s$, which is often the maximum likelihood estimator \cite{lord1980} or a posterior mean estimate by assuming a prior distribution over $\theta_s$ \cite{fox2010bayesian}.
In our work, we use the posterior mean using a Normal$(0,1)$ prior as our reported score.
In the MCEM algorithm, we use the entire posterior distribution to sample $\theta_s$ in the E-step.

\subsection{Feature-based IRT Models and BERT-IRT}

Explanatory IRT models that use item features as in \eqref{eq:mixed_effects} have been used for decades.
Notably, the Log Linear Trait Model (LLTM) of \cite{fischer1973lltm} is an extension of the Rasch model, where the item features predict the item difficulty in a linear model (see also \cite{deboeck2004frameworks}).
Recent extensions of this feature based approach utilize pretrained language models such as BERT \cite{devlin2019bert} to calibrate items from limited response data.
\cite{benedetto2021transformers} predicted item difficulty using a finetuned BERT model from student responses to multiple-choice questions.
\cite{byrd2022predicting} took a similar approach to predicting difficulty and discrimination for 2PL IRT models from natural language questions.
In that work, they constructed linguistically motivated features, such as semantic-ambiguity, as well as contextual embeddings via BERT.
In \cite{reyes2023multiple}, they use RNNs to predict difficulty that is fitted in advance from a Rasch model.
Most of these approaches predict item parameters that have been fit on an existing item bank with many responses as training data.
To train such a model, one needs to already have a large item bank, with hundreds of responses per item, to obtain the necessary training data.
For this reason, we do not include them in our comparisons.

An alternative approach is to train an explanatory IRT model by fitting the non-parametric terms, e.g., $f_d$, using a NNet.
This approach trains a NNet to predict grades from features and the ability $\theta$ while constraining the model to take the IRT form.
BERT-LLTM was proposed in \cite{mccarthy2021jump} to train LLTM models by using BERT embeddings as features along with hand-crafted linguistic features.
More recently, \cite{yancey2024bert} proposed BERT-IRT which fits a 2PL model by constraining the NNet to take the form of \eqref{eq:3PL} and \eqref{eq:mixed_effects}.
In that work, they show that BERT-IRT can accurately train explanatory IRT models with only a handful of test taker responses per item, and apply it to English language proficiency testing.
In that work, they use a linear form for $f_a, f_d$, which we improve on via a fully non-parametric AutoML model.
Furthermore, in that work they use a proxy estimate for the test taker ability, $\tilde \theta_s$, which is constructed from the scores on other item types.
Then they train a model with $\bo x_i, \tilde \theta_s$ as input and the predicted binary grade as output.
Finally, BERT-IRT uses ridge penalties to regularize the random effects parameters $\delta^a, \delta^d$ as well as the coefficients (we perform a coarse grid search, as is outlined in \cite{yancey2024bert}).
We improve on this approach by jointly modeling the ability $\theta_s$ and item parameters $\phi_i$ via the MCEM iteration.

\subsection{Online English Proficiency Testing}

Throughout this study we will analyze data from the Duolingo English Test (DET).
The DET is an online English proficiency test which assesses a test taker's language ability via tasks targeting their speaking, listening, writing, and reading skill.
It contains 14 item types ranging from simple fill-in-the-blank tasks to complex interactive writing and speaking tasks.
In addition to a certified test that can be used for admissions applications to universities and other institutions, there is a practice version that closely mirrors the certified version.
The test consists of several item types that measure different constructs.
All of the data and items evaluated here are from the practice test.

\textbf{Y/N Vocabulary and Vocab-in-Context item types.}  We will focus on the first two item types that are administered in the DET (V8), yes/no vocabulary (Y/N Vocab) and vocabulary in context (ViC).
There are a total of 3290 Y/N Vocab items and 585 ViC items.  
In each test session there are 18 Y/N vocab items and 9 ViC items administered and answered by the test taker.
The test takers are given 5 seconds to respond to a Y/N Vocab item and 20 seconds for ViC items. 
Y/N Vocab asks the test taker if a word is real or not; the words are selected from a pool of real and fake words (the fake words are generated with an RNN).
ViC is a fill in the blank task where the test taker must complete a word that is given in a sentence, i.e., the context.
For example, one Y/N Vocab item asks the test taker if ``newbacal" is a real word, and one ViC item asks to fill in the masked characters in ``I'm sorry for the inter\_\_\_\_\_\_\_, but could you explain that last part again?''.
We display these examples in the supplementary material, Figure \ref{fig:item_examples}.
To see additional examples of these items, one can take the DET practice test for free at \texttt{http://englishtest.duolingo.com}.

In this work, we study the use of AutoIRT for calibrating item parameters for the ViC and Y/N Vocab item types in the DET.
There are three primary use cases for calibration, cold-start and jump-start piloting of new items, and warm-start recalibration of the entire item bank.
\begin{enumerate}
    \item \emph{Cold-start:} new pilot items are added to the test with no response data.  In this scenario, we are only able to use item features to calibrate the item parameters.  However, we have an operational item bank with many responses at our disposal.  In this setting, we will re-calibrate the operational item bank, as well as calibrate the pilot items.
    \item \emph{Jump-start:} we gather a limited number of responses to new pilot items in a piloting phase.  Similar to the cold-start setting, we have an operational item bank with many responses at our disposal and we re-calibrate the operational as well as calibrate the pilot items.
    \item \emph{Warm-start:} we want to re-calibrate an operational item bank in order to make the new item parameters reflective of the most recent response data.  This can be triggered by distribution shift due to changes in the user interface of the test, the test taker population, or improved test prep materials.
\end{enumerate}

\section{Method}

We separately model the ViC and Y/N Vocab item parameters and ability, so that each item type has its own associated $\theta_s$. 
Throughout we will jointly estimate $\theta_s$ and $\phi_i$ via the MCEM algorithm below.
We will compare against BERT-IRT as described in \cite{yancey2024bert} which has access to the same features as our model.

\subsection{Item Features}



For each item, we compute a set of features based on the item content. For features based on analysis of the word's use in a corpus, we use the Corpus of Contemporary American English (COCA) \cite{davies2008word}.

In the case of Y/N Vocab, the item consists of a single word. Some features are evident from the item in isolation: a binary indicator is\_real for the answer and the length in characters. For each CEFR level \cite{councilofeurope2001}, we include binary indicators of whether the word appears in a CEFR level-specific wordlist.
We also use several corpus frequency features: the log frequency of the word, the log frequency-rank of the word, and the proportion of occurrences of the word that are capitalized. Finally, we include features based on n-gram frequencies computed from COCA: for each $n$ from 1 to 4, we consider the frequency of the prefix and suffix of size $n$; number of $n$-grams in the word with frequency over a fixed threshold; and minimum, average, and maximum frequencies of $n$-grams within the word.

For ViC, we use some simple surface features: the number of missing characters and the proportion of vowels in the missing portion. Other features are more focused on the contexts in which the word is commonly used, including the log frequency of the target word in COCA as a whole and in each of the 8 sub-corpora of COCA, and the log \textit{document} frequency of the word. We include BERT embeddings of the masked passage---we reduce the dimensions to 10 using PCA.  We include some features focusing on the sentence, namely the \textit{average} log frequency in COCA over all words in the sentence, and the position of the damaged word within the sentence, normalized by the sentence length. Finally, we compute the conditional probability of the correct word, given the visible letters, derived using unigram COCA frequencies of words consistent with the visible prefix and the length.

\subsection{Monte Carlo Expectation Maximization}

The expectation maximization (EM) algorithm will consider the ability $\theta_s$ to be a stochastic latent quantity with a known prior, and the item parameters $\phi_i$ are non-stochastic latent quantities.
In the E-step, we will update the posterior $\pi( \theta | \{G_{i,s}\}_{i \in \cl I_s}, \mathbf{a}, \mathbf{d})$ based on the grades.
Maintaining and updating the posterior is simplified by the fact that $\theta$ is unidimensional, and we use a discretization over a grid of $\theta$ as an approximation.

We will sample $\theta$s from this posterior and then treat it as fixed in our M-step.
The products of our M-step are fitted item parameters given fixed ability, $\bf \theta$. We fit the item parameters $\phi_i$ by minimizing the negative binary log-likelihood,
\begin{equation}
\begin{split}
    \ell(\bo{\phi} | \bo{\theta}) = & \sum_{(i,s) \in \cl R} G_{i,s} \log p(\theta_s, \phi_i) \\
    & + (1 - G_{i,s}) \log \left( 1 - p(\theta_s, \phi_i)\right)
\end{split}
\end{equation}
where $\theta_s \sim \pi(\theta | \bo G_s, \bo \phi)$ independently.
This is unbiased for the $\bo \theta$ dependent term in the evidence lower bound (ELBO) \cite{bishop2006pattern}.

In the E-step of the MCEM algorithm, we will draw a single ability parameter for each session from the posterior distribution, which constitute the fixed $\bf \theta$ in the M-step.
We initialize $\theta_s$ for each session $s = 1, \ldots, S$ --- typically this is a weighted average of scores for other tasks in the test.

\begin{algorithm}
\caption{AutoIRT Calibration}
\begin{algorithmic}
\STATE Start with initial ability parameters $\bf \theta$.
\FOR{each EM iteration}
    \STATE \textbf{M-step}:
    \STATE Let $Z_{i,s} = (\theta_s, X_i)$ and $G_{i,s} \in \{0,1\}$ be the grade for item $i$ in session $s$.
    \STATE Fit an AutoML classifier to predict $G_{i,s}$ from $Z_{i,s}$, let $\hat{p}(\theta, X_i)$ be the predicted grade.
    \STATE Fit the item parameters $\bo \phi$ by minimizing the least squares objective \eqref{eq:leastsquares}.
    \STATE \textbf{E-step} (skip last iteration):
    \FOR{each session $s$}
        \STATE Approximate the posterior distribution: $\pi(\theta_s | \{G_{i,s}, \phi_i\}_{i \in \cl I_s})$.
        \STATE Draw $\theta_s \sim \pi( . | \{G_{i,s}, \phi_i\}_{i \in \cl I_s})$.
    \ENDFOR
\ENDFOR
\end{algorithmic}
\label{alg:autoirt}
\end{algorithm}

The final results are estimates of the item parameters, $\phi_i$.
One advantage of this approach is that the $\theta$ posterior is used for both scoring and calibration.
Note that this is an empirical Bayesian approach because it still uses maximum marginal likelihood to estimate the item parameters.
Alternatives are fully Bayesian approaches that are typically based on MCMC samplers, or joint estimation of $\theta$ and $\phi$.
However, there is no literature on using AutoML with either of these frameworks.

\subsection{AutoIRT M-step}

In order to leverage the flexibility and model performance of AutoML, we will first fit a grade classifier using $\bo \theta$ and $d$-dimensional item features $\bo x_i \in \bb R^d$ as input.
In our experiments, we will use a stacked ensemble of random forests, lightGBM, XGBoost, and CATBoost via the AutoGluon-tabular python package \cite{erickson2020autogluon}.
AutoGluon was chosen due to its consistently good performance in benchmarks on tabular data \cite{gijsbers2024amlb}.
We use tabular models only because we have already performed feature engineering, including using BERT embeddings.
However, AutoML tools support multimodal input and in the future we hope that AutoIRT will be used directly on item content.
We also pass the item id as a feature to the AutoML predictor, similar to the use of random effects terms in \eqref{eq:mixed_effects}.
Let $\hat p(\theta, \bo x_i)$ denote the AutoML predicted probability of correct response for a test taker with ability parameter $\theta$ and item with features $\bo x_i$.

From the trained AutoML predictor we want to extract the more interpretable, trained IRT model \eqref{eq:3PL}.
We do this by projecting the AutoML model onto the closest IRT model in the least squares sense.
Specifically, we select the item parameters that minimize the following loss,
\begin{equation}
    \label{eq:leastsquares}
    L(\bo \phi) = \sum_{i \in \cl R} \sum_{\theta \in \Theta} (p(\theta, \phi_i) - \hat p(\theta, \bo x_i))^2,
\end{equation}
where $\Theta$ is a regular grid of $\theta$s, such as $(-3, -2.9, \ldots, 2.9, 3)$.
The main requirement on the grid $\Theta$ is that it contains the majority of the probability mass of the distribution of $\theta$s.

\subsection{Evaluating Calibrated Item Parameters}

Algorithm \ref{alg:autoirt} produces the following key artifacts: posterior distributions of $\theta_s$ for the training sessions, a trained AutoML predictor from the last M-step, and IRT parameters $\hat \phi_i$ for operational items $i$ (items in the training set).
In the cold-start setting, there are pilot items for which we do not have any training responses.
In order to obtain $\hat \phi_i$ for a pilot item, $i$, we use the final AutoML predictor to compute $\hat p(\theta, \bo x_i)$ over $\theta \in \Theta$, and then obtain $\phi_i$ by minimizing \eqref{eq:leastsquares}.

Evaluating IRT models on a test set is complicated by the fact that $\theta$ is unknown for any test sessions.
We circumvent this by scoring each session in a test session, i.e., using the predicted IRT parameters ($\hat \phi_i$) to obtain the posterior mean for that given session,
\begin{equation}
    \hat \theta_s = \int_{-\infty}^\infty \theta \cdot \pi(\theta | \{G_{s,i}, \hat \phi_i \}_{i \in \cl I_s}) \textrm{d} \theta.
\end{equation}
Armed with this estimate for a test session, we can then evaluate the negative log-likelihood (binary cross-entropy) for that session,
\begin{equation*}
    \sum_{i \in \cl I_s} G_{i,s} \log p(\hat \theta_s, \hat \phi_i) + (1 - G_{i,s}) \log \left( 1 - p(\hat \theta_s, \hat \phi_i)\right),
\end{equation*}
giving us our test loss ($\cl I_s$ are the items administered to that session, $s$).
Similarly, we compare the item mean grade correlation (Pearson and Spearman), which is the correlation between the mean grade and mean predicted grade for each response when grouped by item.

Another common performance measure for scores is the \emph{retest-reliability} (RR), namely the Pearson correlation between scores for repeat tests --- when the same user takes the test twice in succession.
In the classical test theory model, it is assumed that the observed score is an unobserved ``true'' score plus a noise component.
Given some modeling assumptions, the standard error of measurement (SEM), the standard error for the score as an estimate of the true score, is related to the RR through the formula,
\begin{equation}
\label{eq:sem}
    S_E = S_X \sqrt{1 - RR},
\end{equation}
where $S_X$ is the standard deviation of the scores for the entire population of test takers, and $S_E$ is the standard deviation of the error term, \cite{lord1968statistical}.
Hence an increase of $RR$ from 0.5 to 0.6 results in a decrease in SEM of $11.8\%$.
There are many measures of reliability, and RR is a practical measure for adaptive tests like the DET.

Offline evaluation suffers from the fact that the items are administered according to an existing online policy.
All of the aforementioned measures are based on averages over the sampled items and if the online policy biases the sample toward certain items-session pairs then this bias can influence these measures.
One concrete byproduct of this is that since $\hat \theta_s$ is based on the performance of the test taker, then this is associated with the items selected.
The result is that the item mean grade correlation is typically positive even when the item parameters are constant for all items.

For online experiments it is important to take into consideration the fact that the fitted item parameters have an impact on the test administration policy.
For the DET, these vocab items are selected such that the item difficulty parameter, $d_i$, falls within the bulk of the posterior for $\theta_s$, with preference given to high discrimination items.
In this work, we withhold more details about the administration policy used, to protect intellectual property.
The typical result of additionally using the fitted item parameters is that the RR is higher than in the offline analysis.



\section{Experimental Results}

\subsection{Simulation study}

In order to demonstrate the ability of AutoIRT to estimate item parameters that are complex functions of their features, we simulate data from a 3PL model.
In the simulation, there are two features, $X_{i,1}, X_{i,2} \sim \mathcal{U}(-10, 10)$ iid for each item.
We generate $\theta_j \sim \mathcal{N}(0, 2.5)$ for each $j = 1, \ldots, N_{sess}$ (the total number of sessions), and $a_i, d_i, c_i$ (slope, difficulty, and chance) for $i = 1, \ldots, N_{item}$ (the total number of items).
The chance parameter is set to $c_i = 0.25$, and $a_i, d_i$ can be seen in Fig.~\ref{fig:simulation_details} (details are given in the Supplement).
The final grades follow the 3PL model \eqref{eq:3PL}.

\begin{figure}
    \centering
    \includegraphics[width=\linewidth]{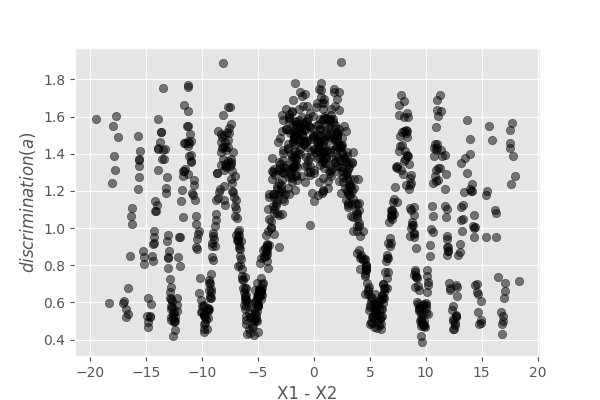}
    \includegraphics[width=\linewidth]{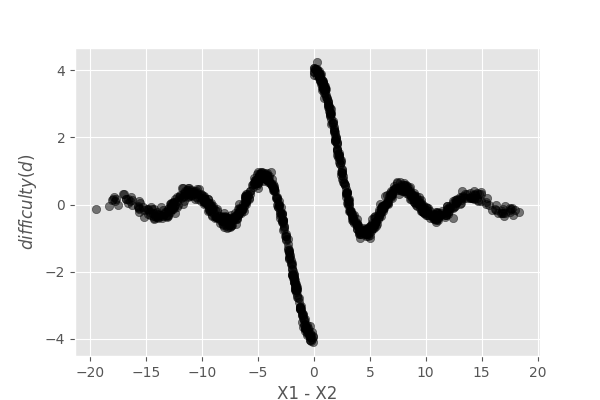}
    \caption{Simulation of item parameters, discrimination (a, top) and difficulty (d, bottom) as a function of the item features.  Each dot constitutes an item.}
    \label{fig:simulation_details}
\end{figure}

Throughout, we evaluate the loss as the mean negative log-likelihood.
We simulate the warm-start setting via a test set by holding out sessions but using the same item bank as the training set, and the cold-start setting by creating a held-out pilot set of items and sessions so that the training set and hold-out set do not share any items or sessions.
We also look at the item mean grade correlation (Pearson).
Finally, because this is a simulation and we know the true $\theta$ for a given session, we can look at the correlation between the estimated $\theta$, or score, (which is the posterior mean for a session) and the true $\theta$.
The parameter settings are as follows: the number of items is 100, 400, and 1600; the number of sessions is 1250, 2500, 10000, 40000, and 160000; the number of items administered per session is 10; the random effect standard deviation is 0.1; the number of steps is 4; the number of out-of-sample (held-out) items is 1000; and the number of test sessions is 100,000.

The main conclusions are that we see the expected pattern of increased performance in the warm-start setting when we increase the number of sessions, and in the cold-start setting, the quality of the item parameters are dramatically impacted by the total number of items in the bank.
First, we find that within 4 steps of the EM algorithm, the training losses are roughly constant, which is important because each M-step requires an AutoML fit which can be computationally expensive (see details in the Supplement).
Furthermore, after the first M-step, the AutoML training loss is lower than the fitted 3PL model in the M-step, but for the vast majority of simulations this trend reverses at the subsequent steps.
The fact that the 3PL model gives lower training error is an important sanity check because the data follows this model.
Second, in the warm-start setting, we are able to achieve a mean grade-predicted probability correlation above 0.95 with 10000 sessions (even for 1600 items).
The resulting session scores are well correlated with the latent ability $\theta$ for all of the simulation settings (correlation is between 0.84 and 0.87), which indicates that while the predictive performance is improved by observing more sessions, the score accuracy is not as sensitive.
Third, in the cold-start setting, we see that the predictive performance is dramatically improved by the number of items.
For 10000 sessions, the mean grade-predicted probability correlation jumps from $0.58$ to $0.83$ when we increase the number of training items from 400 to 1600.

\begin{figure}
    \centering
    \includegraphics[width=\linewidth]{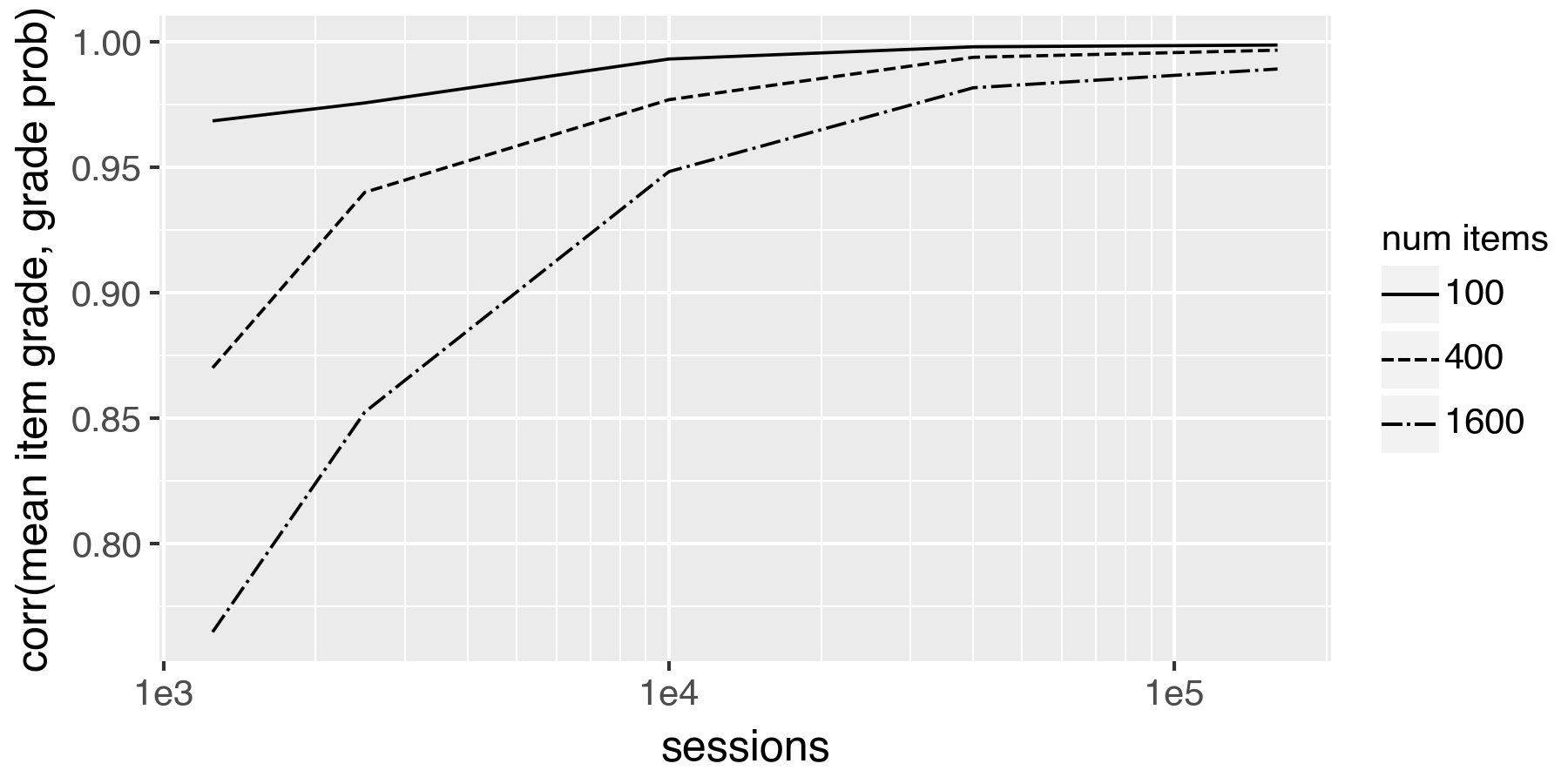}
    \includegraphics[width=\linewidth]{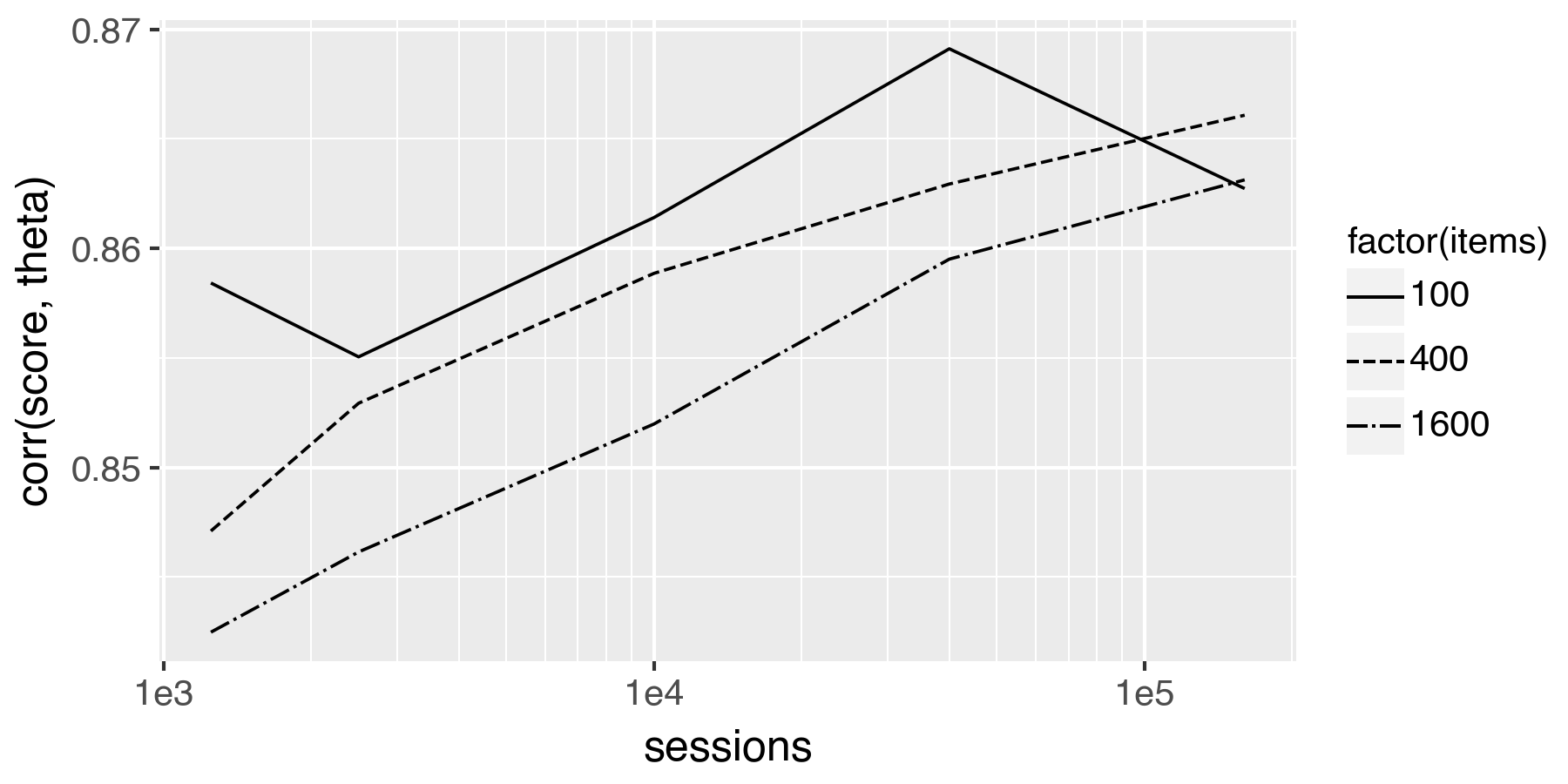}
    \caption{Warm-start evaluation on simulated held-out sessions: the correlation between the average grade across items and the average predicted grades (top) and the correlation between the score and latent ability $\theta$ (bottom).}
    \label{fig:sim-test}
\end{figure}

\begin{figure}
    \centering
    \includegraphics[width=\linewidth]{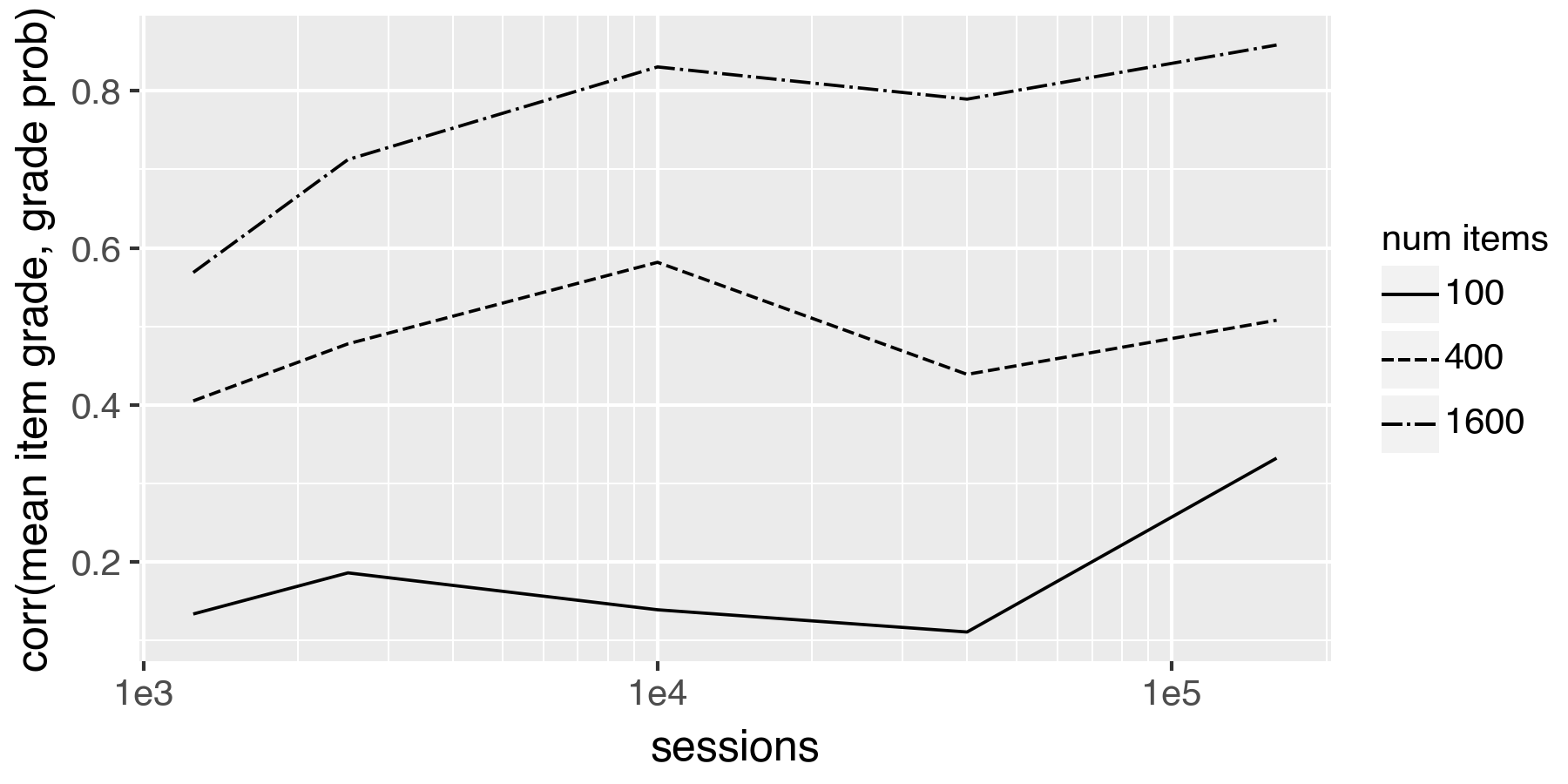}
    \includegraphics[width=\linewidth]{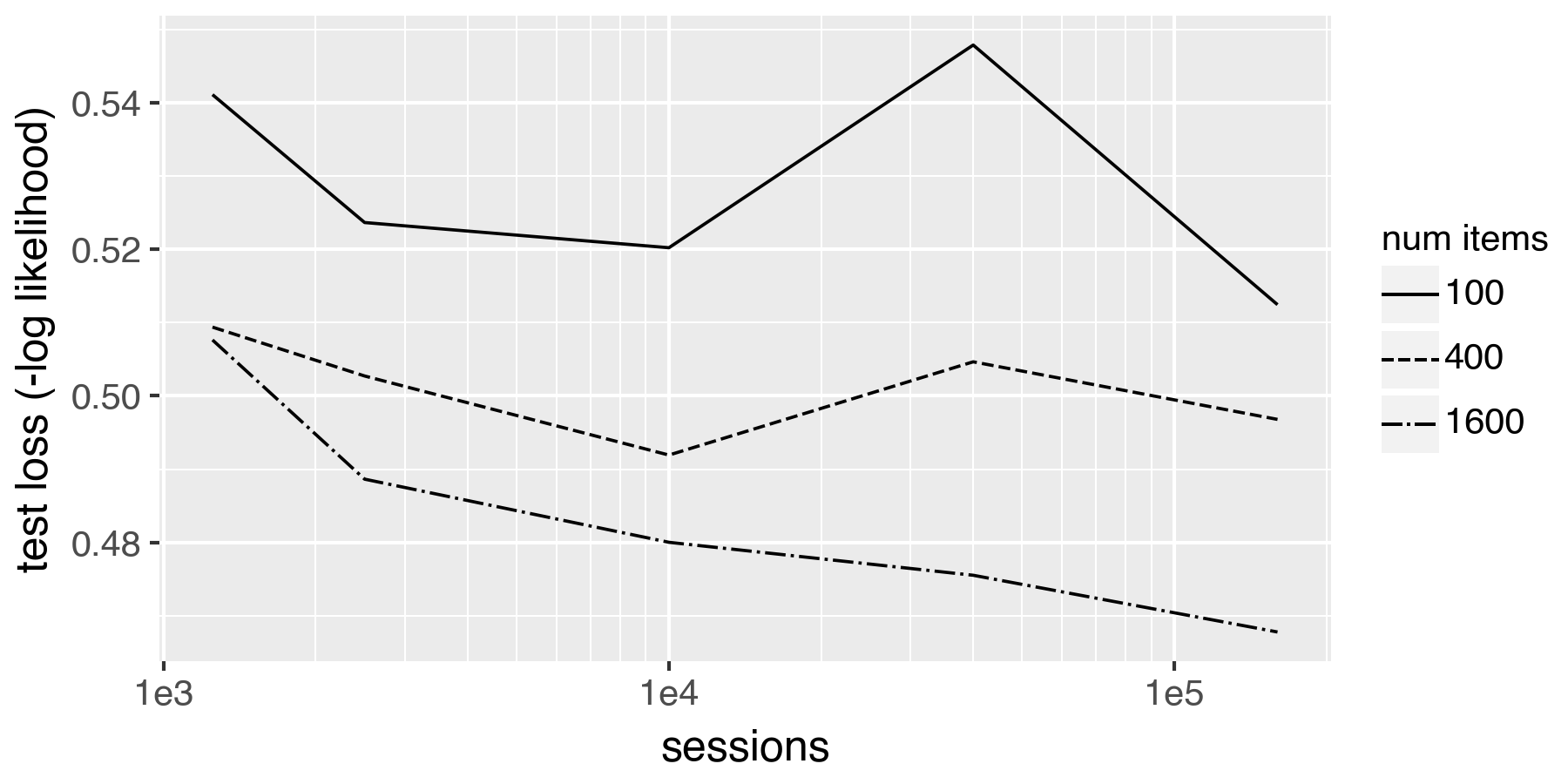}
    \caption{Cold-start evaluation on simulation held-out items: the correlation between the average grade across items and the average predicted grades (top) and the test loss for those items and sessions that are held-out (bottom).}
    \label{fig:sim-oos}
\end{figure}

\subsection{Offline calibration analysis for DET}

We will look at offline evaluations of DET practice test responses for Y/N Vocab and ViC items.
We compare the performance of these methods on practice test taker data between 2024-04-01 and 2024-06-18.
We compare the non-explanatory 2PL model (labeled IRT), the BERT-IRT 2PL model, and AutoIRT.
For Y/N vocab, we set $c = 0.25$ in AutoIRT to account for the possibility that the test taker simply guesses the answer, which is possible with multiple choice answers (the specific value was determined through an initial exploratory analysis).
For ViC, $c=0$ because it is less likely that the test taker will simply guess the characters needed to complete the word.
We split the data in multiple ways to simulate the warm-start, jump-start, and cold-start scenarios outlined below.

\begin{enumerate}
    \item Cold start train-test split: We select a split date (2024-05-22) and randomly hold out a set of pilot items (50-50\% split of pilot and operational items).  The training set consists of all responses for operational items prior to the split date, and the test set consist of all responses past the split date for all items.
    \item Jump start train-test split: We select the same time and item splits as in the cold-start scenario.  The training set consists of all operational item responses prior to the split date as well as the first R responses for the pilot items after the split date (R=20, 40, 80).  The test set is the remainder of responses after the split date (``Jump 20'' means R=20).
    \item Warm start train-test split: We split the responses into prior to the split date and after the split date.  We use split dates: 2024-05-08 and 2024-05-15.
\end{enumerate}

\begin{table}[ht]
\centering
\begin{tabular}{lcccc}
\hline
\textbf{Split} & \textbf{Method} & \textbf{Loss} & \textbf{Pears.} & \textbf{Spear.} \\
\hline
Cold & BERT-IRT & 0.462 & \textbf{0.814} & \textbf{0.831} \\
Cold & IRT & 0.475 & 0.718 & 0.792 \\
Cold & AutoIRT & \textbf{0.456} & 0.785 & 0.822 \\
Jump 20 & BERT-IRT & 0.453 & 0.901 & 0.916 \\
Jump 20 & IRT & 0.460 & 0.855 & 0.903 \\
Jump 20 & AutoIRT & \textbf{0.428} & \textbf{0.936} & \textbf{0.944} \\
Jump 40 & BERT-IRT & 0.448 & 0.936 & 0.944 \\
Jump 40 & IRT & 0.453 & 0.914 & 0.938 \\
Jump 40 & AutoIRT & \textbf{0.426} & \textbf{0.955} & \textbf{0.962} \\
Jump 80 & BERT-IRT & 0.445 & 0.961 & 0.963 \\
Jump 80 & IRT & 0.447 & 0.952 & 0.959 \\
Jump 80 & AutoIRT & \textbf{0.423} & \textbf{0.971} & \textbf{0.973} \\
Warm 05-08 & BERT-IRT & 0.449 & 0.982 & 0.982 \\
Warm 05-08 & IRT & 0.449 & 0.986 & 0.985 \\
Warm 05-08 & AutoIRT & \textbf{0.431} & \textbf{0.987} & \textbf{0.991} \\
Warm 05-15 & BERT-IRT & 0.441 & 0.980 & 0.980 \\
Warm 05-15 & IRT & 0.440 & 0.983 & 0.983 \\
Warm 05-15 & AutoIRT & \textbf{0.423} & \textbf{0.987} & \textbf{0.989} \\
\hline
\end{tabular}
\caption{Y/N Vocab offline analysis. We report binary cross-entropy loss and item mean grade correlation (Pearson and Spearman) for the cold-start, jump-start, and warm-start settings.}
\label{table:stv}
\end{table}

\begin{table}[ht]
\centering
\begin{tabular}{lcccc}
\hline
\textbf{Split} & \textbf{Method} & \textbf{Loss} & \textbf{Pears.} & \textbf{Spear.} \\
\hline
Cold & BERT-IRT & \textbf{0.425} & \textbf{0.781} & \textbf{0.827} \\
Cold & IRT & 0.441 & 0.683 & 0.679 \\
Cold & AutoIRT & 0.435 & 0.742 & 0.786 \\
Jump 20 & BERT-IRT & 0.404 & 0.890 & 0.923 \\
Jump 20 & IRT & 0.412 & 0.847 & 0.899 \\
Jump 20 & AutoIRT & \textbf{0.381} & \textbf{0.954} & \textbf{0.959}\\
Jump 40 & BERT-IRT & 0.391 & 0.943 & 0.952 \\
Jump 40 & IRT & 0.397 & 0.918 & 0.940 \\
Jump 40 & AutoIRT & \textbf{0.374} & \textbf{0.978} & \textbf{0.979} \\
Jump 80 & BERT-IRT & 0.382 & 0.973 & 0.975 \\
Jump 80 & IRT & 0.385 & 0.963 & 0.967 \\
Jump 80 & AutoIRT & \textbf{0.371} & \textbf{0.989} & \textbf{0.989} \\
Warm 05-08 & BERT-IRT & 0.368 & 0.994 & 0.989 \\
Warm 05-08 & IRT & 0.369 & 0.992 & 0.985 \\
Warm 05-08 & AutoIRT & \textbf{0.359} & \textbf{0.998} & \textbf{0.997} \\
Warm 05-15 & BERT-IRT & 0.365 & 0.995 & 0.990 \\
Warm 05-15 & IRT & 0.365 & 0.994 & 0.987 \\
Warm 05-15 & AutoIRT & \textbf{0.355} & \textbf{0.998} & \textbf{0.997} \\
\hline
\end{tabular}
\caption{ViC offline analysis.  We report binary cross-entropy loss and item mean grade correlation (Pearson and Spearman) for the cold-start, jump-start, and warm-start settings.}
\label{table:vic}
\end{table}

\begin{table}[ht]
\centering
\begin{tabular}{lccc}
\hline
\textbf{Split} & \textbf{BERT-IRT} & \textbf{IRT} & \textbf{AutoIRT} \\
\hline
Cold & \textbf{0.522} & 0.518 & 0.495 \\
Jump 20 & 0.527 & 0.525 & \textbf{0.535} \\
Jump 40 & 0.528 & 0.527 & \textbf{0.537} \\
Jump 80 & 0.528 & 0.529 & \textbf{0.538} \\
Warm 05-08 & 0.533 & 0.536 & \textbf{0.541} \\
Warm 05-15 & 0.536 & 0.539 & \textbf{0.545} \\
\hline
\end{tabular}
\caption{Y/N Vocab offline retest reliability}
\label{table:stv_retest}
\end{table}

\begin{table}[ht]
\centering
\begin{tabular}{lccc}
\hline
\textbf{Split} & \textbf{BERT-IRT} & \textbf{IRT} & \textbf{AutoIRT} \\
\hline
Cold & 0.498 & 0.485 & \textbf{0.526} \\
Jump 20 & 0.510 & 0.503 & \textbf{0.546} \\
Jump 40 & 0.518 & 0.513 & \textbf{0.550} \\
Jump 80 & 0.523 & 0.520 & \textbf{0.551} \\
Warm 05-08 & 0.541 & 0.540 & \textbf{0.565} \\
Warm 05-15 & 0.536 & 0.536 & \textbf{0.561} \\
\hline
\end{tabular}
\caption{VIC offline retest reliability}
\label{table:vic_retest}
\end{table}

We compare the results for our three methods in the 5 different test settings.
In none of the test settings do non-explanatory IRT models outperform their feature-based counterparts, BERT-IRT and AutoIRT.
In Table \ref{table:stv}, we see that in the cold-start setting BERT-IRT has better correlation measures for both item types, but this trend reverses in the jump-start setting.

For Y/N Vocab, the binary cross-entropy loss is significantly better for AutoIRT relative to BERT-IRT and IRT.
This is likely due to the fact that additionally learning $\theta$ in the E-step leads to less biased predicted probabilities.
This effect is more dramatic when there is a non-zero chance parameter, and corresponding chance that the test taker can guess the right answer.

\begin{figure}
    \centering
    \includegraphics[width=\linewidth]{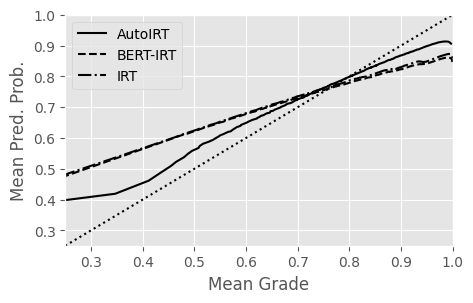}
    \caption{Calibration comparison of three methods for Y/N Vocab in Jump-start 20 setting; the mean grades and pred.~prob. are averages grouped by $\theta$ \%ile.  These two means should match for a well calibrated model.}
    \label{fig:calib-comp}
\end{figure}

The cross-entropy loss reflects the accuracy of the predicted probabilities in a way that the correlation metrics will not.
We can see in Figure \ref{fig:calib-comp} that the mean grades match more closely the mean predicted probabilities when we average them grouped by $\theta$ (binned by percentile).
We have observed that using cross-entropy validation loss in the AutoML tool is important to keep your model well calibrated overall.  
Using accuracy as the validation metric tends to lead to a greater mismatch between the mean predicted probabilities and the mean grades.

\subsection{Offline and online reliability analysis for DET}

Reliability as a measure gives unique insights into the performance of a calibrated item bank.
According to \eqref{eq:sem}, an RR of $0.5$ results in a $30.3\%$ reduction of standard error of measurement from the baseline (no measurements).
All 18 Y/N Vocab items take at most 90 seconds to complete, while all 9 ViC items take 180 seconds, which is a small price to pay for this improvement in reliability.
Offline RR is higher for AutoIRT relative to BERT-IRT in all scenarios except cold-starting Y/N Vocab.
This is consistent with the observation that BERT-IRT is most competitive in the cold-start setting.

\begin{table}
\begin{tabular}{@{}l@{~}llrrr@{}}
\hline
\textbf{Experiment} & \textbf{Cond} & \textbf{N sess} & \textbf{ViC} &  \textbf{Y/N} &  \textbf{Overall} \\
\hline
01-26 E1 & C & & & $^1$0.712 & 0.812 \\
         & T1 & 15782 & 0.511 & 0.655 & 0.817 \\
         & T2 & 15796 & 0.507 & 0.647 & 0.849 \\
01-31 E2 & C & & & 0.714 & 0.794 \\
         & T2 & 20266 & 0.515 & 0.594 & 0.830 \\
         & T3 & 19724 & 0.505 & 0.586 & 0.826 \\
02-12 E3 & C & & & 0.709 & 0.821 \\
         & T2 & 22248 & 0.486 & 0.558 & 0.833 \\
         & T4 & 11044 & 0.486 & 0.538 & 0.829 \\
02-16 E4 & C & & & 0.712 & 0.804 \\
         & T5 & 7799 & $^2$0.634 & 0.526 & 0.771 \\
         & T6 & 15417 & 0.640 & 0.567 & 0.814 \\
03-01 Launch & T5 & & 0.633 & $^3$0.572 & $^4$0.824 \\
\hline
\end{tabular}
\caption{Retest reliabilities for relevant DET practice test experiments.  The control condition (C) is DET V7 which contains 4 sets of 18 Y/N Vocab words, all treatment conditions have 18 Y/N Vocab items, followed by 8 ViC items.}
\label{tab:online_reliabilities}
\end{table}

Between 2024-01-01 and 2024-03-01, several experiments were run on the DET practice test culminating in the launch of ViC and Y/N Vocab on the practice test (prior to launch on the certified DET).
Those listed in the Table \ref{tab:online_reliabilities} are experiments that tested the reliability of both ViC and Y/N Vocab and how they impact the overall RR of the final overall score of the practice test.
All treatment conditions (T1-T6) use vocab items that are calibrated with AutoIRT.
Initially, the only calibration data for the ViC item type was a limited piloting stage that occurred at the end of the practice test for a short period in 2023; this was used in E1-E3, T1-T4.
For experiment E4 (T5, T6) and the launch, we calibrated ViC using all of the practice test data gathered thus far.
Most of these treatment arms tested changes to the DET administration, and so most of the differences between the treatment arms are not salient to this study.
All of the treatment arms listed here use the same Y/N Vocab calibration from practice test data using AutoIRT.
The control condition is the production DET practice test (V7) which does not contain ViC and has Y/N Vocab items in an 18 word card format and 4 such items.
Thus, the control condition has 4 times the number of Y/N vocab words.
To summarize our experimental conditions:
\begin{enumerate}
    \item[\bf C] Control, 4 cards of 18 Y/N Vocab words, Y/N Vocab is calibrated with historical data using 2PL non-explanatory IRT, no ViC.
    \item[\bf T1-4] Treatment conditions with ViC and Y/N Vocab, ViC is calibrated with a limited pilot, and Y/N Vocab is calibrated with practice test responses.  The differences between these are test changes that do not impact the vocab items.
    \item[\bf T5-6] Same as T1-T4 except that ViC is recalibrated with more recent PT data.
\end{enumerate}

Naturally, launching Y/N Vocab in the single word format will reduce its reliability since control has 4 times the number of Y/N words.
We see that our Y/N vocab RR in the control condition, hovers around 0.7 ($^1$ in Table \ref{tab:online_reliabilities}), and according to classical test theory, we can expect that without improvements in calibration this will be reduced to 0.495.
We find that in fact the improvements in calibration results in Y/N vocab RR of 0.572 at launch resulting in a 15.5\% increase over baseline ($^3$ in Table \ref{tab:online_reliabilities}).
There are several deviations from this result likely due to population shifts in the DET at the time of testing, but the treatment conditions have average 0.584 which is close to the launch Y/N Vocab RR.
The DET overall score is a predetermined linear combination of the scores for all item types in the DET (9 in total).
In general, the RR is increased in the treatment and launch conditions, and the switch to the new vocab item types only results in a 30 second longer test ($^4$ in Table \ref{tab:online_reliabilities}).
Finally, recalibrating ViC with the practice test data collected prior to E4 resulted in a dramatic improvement in reliability ($^2$ in Table \ref{tab:online_reliabilities}).
The condition T5 was chosen to be the final launch candidate, and the drop in reliability in E4 was deemed to be spurious.
This was confirmed after launch where the final reliability metrics were much improved.



\section{Conclusions}

There are two key innovations presented here: (1) the MCEM iteration jointly learns the session level abilities, $\theta_s$, and the item parameters, $\phi_i$; and (2) we use AutoML for IRT modeling.
From our simulation study, we see that the number of items in the existing item bank drives the performance in the cold-start and jump-start settings.
This suggests that as our item bank grows we are more able to cold-start pilot items which will further accelerate the growth of the item bank.
Furthermore, AutoIRT significantly improved the cross-entropy loss, which reflects the improved output probability bias.
One key advantage of our method is that there is no need for hyperparameter tuning and can support multi-modal input.
However, we do not take advantage of the multi-modal support and we still use hand-crafted item features.
AutoIRT needs to be applied to more complex multi-modal item types for it to reach its full potential.

\bibliography{autoirt}

\newpage

\section*{Supplementary Material for: \\ \emph{AutoIRT: Calibrating Item Response Theory Models with Automated Machine Learning}}
\addcontentsline{toc}{section}{Supplementary Material}

\appendix

\section{Simulation details}
\label{sec:simulation_details}

The simulated 3PL model is given by the following equations.
It generates $\theta_j$ for each $j = 1, \ldots, N_{sess}$ (the total number of sessions), and $a_i, d_i, c_i$ (slope, difficulty, and chance) for $i = 1, \ldots, N_{item}$ (the total number of items).
We set the random effects standard deviation to be $\sigma_{rand} = 0.1$.

\begin{equation}
\theta_j \sim \mathcal{N}(0, 2.5)
\end{equation}
for \( j = 1, 2, \ldots, N_{\text{sess}} \)

\begin{equation}
X_{i,1} \sim \mathcal{U}(-10, 10) \quad \text{and} \quad X_{i,2} \sim \mathcal{U}(-10, 10)
\end{equation}
for \( i = 1, 2, \ldots, N_{\text{item}} \)

\begin{equation}
\bo x_i = (x_{i,1}, x_{i,2})
\end{equation}

\begin{equation}
Z_i = X_{i,1} - X_{i,2}
\end{equation}
for \( i = 1, 2, \ldots, N_{\text{item}} \)

\begin{equation}
d_{\text{mean}}^i = \frac{4 \sin(Z_i)}{|Z_i|}
\end{equation}
for \( i = 1, 2, \ldots, N_{\text{item}} \)

\begin{equation}
a_{\text{mean}}^i = 0.5 \cos(0.1 Z_i^2) + 1.0
\end{equation}
for \( i = 1, 2, \ldots, N_{\text{item}} \)

\begin{equation}
d_i = d_{\text{mean}}^i + \delta_d^i \quad \text{where} \quad \delta_d^i \sim \mathcal{N}(0, \sigma_{\text{rand}})
\end{equation}
for \( i = 1, 2, \ldots, N_{\text{item}} \)

\begin{equation}
a_i = \exp(\log(a_{\text{mean}}^i) + \delta_a^i) \quad \text{where} \quad \delta_a^i \sim \mathcal{N}(0, \sigma_{\text{rand}})
\end{equation}
for \( i = 1, 2, \ldots, N_{\text{item}} \)

\begin{equation}
c_i = 0.25 \quad \text{(constant for all items)}
\end{equation}
for \( i = 1, 2, \ldots, N_{\text{item}} \)
The resulting item parameters are $a_i, d_i, c_i$ as the slope, difficulty, and chance parameters respectively.

The following are all of the inputs for the simulations.  Each combination of number of items and number of sessions are used.
\begin{itemize}
    \item Number of items: 100, 400, and 1600
    \item Number of sessions: 1250, 2500, 10000, 40000, and 160000
    \item Number of items administered per session: 10
    \item Random effect standard deviation: 0.1
    \item Number of steps: 4
    \item Number of out-of-sample items: 1000
    \item Number of test sessions: 100000
\end{itemize}

\begin{figure*}
    \centering
    \includegraphics[width=.8\linewidth]{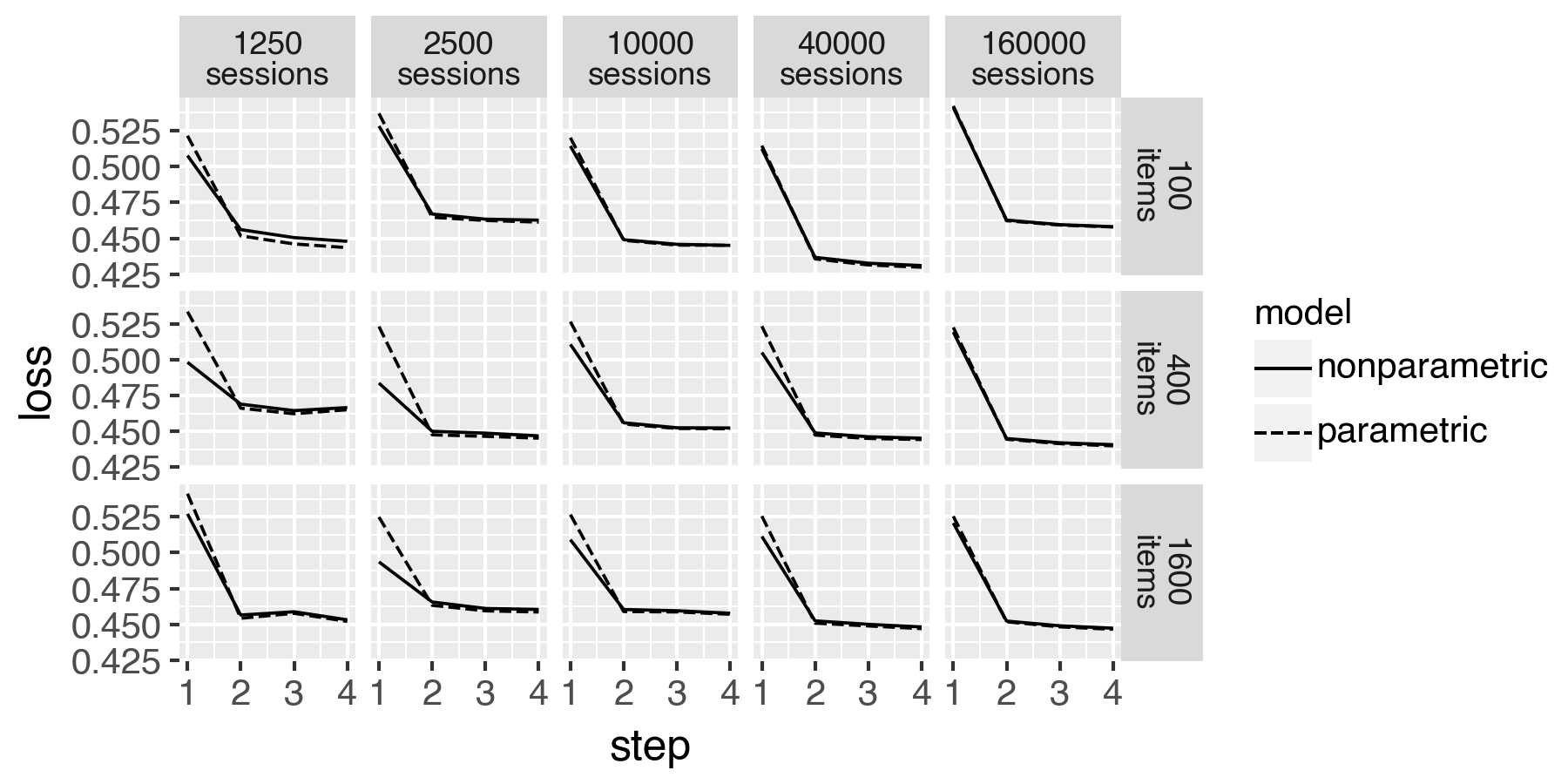}
    \caption{The training loss for based on the EM iterations.  We report all simulation settings by considering the total number of sessions and items used for training.  The non-parametric model is the training loss for model returned from the AutoML step, the parametric model is the training loss for the 3PL models that is fitted to the AutoML output in the M-step.}
    \label{fig:sim-training-losses}
\end{figure*}

\begin{figure*}
    \centering
    \includegraphics[width=0.49\linewidth]{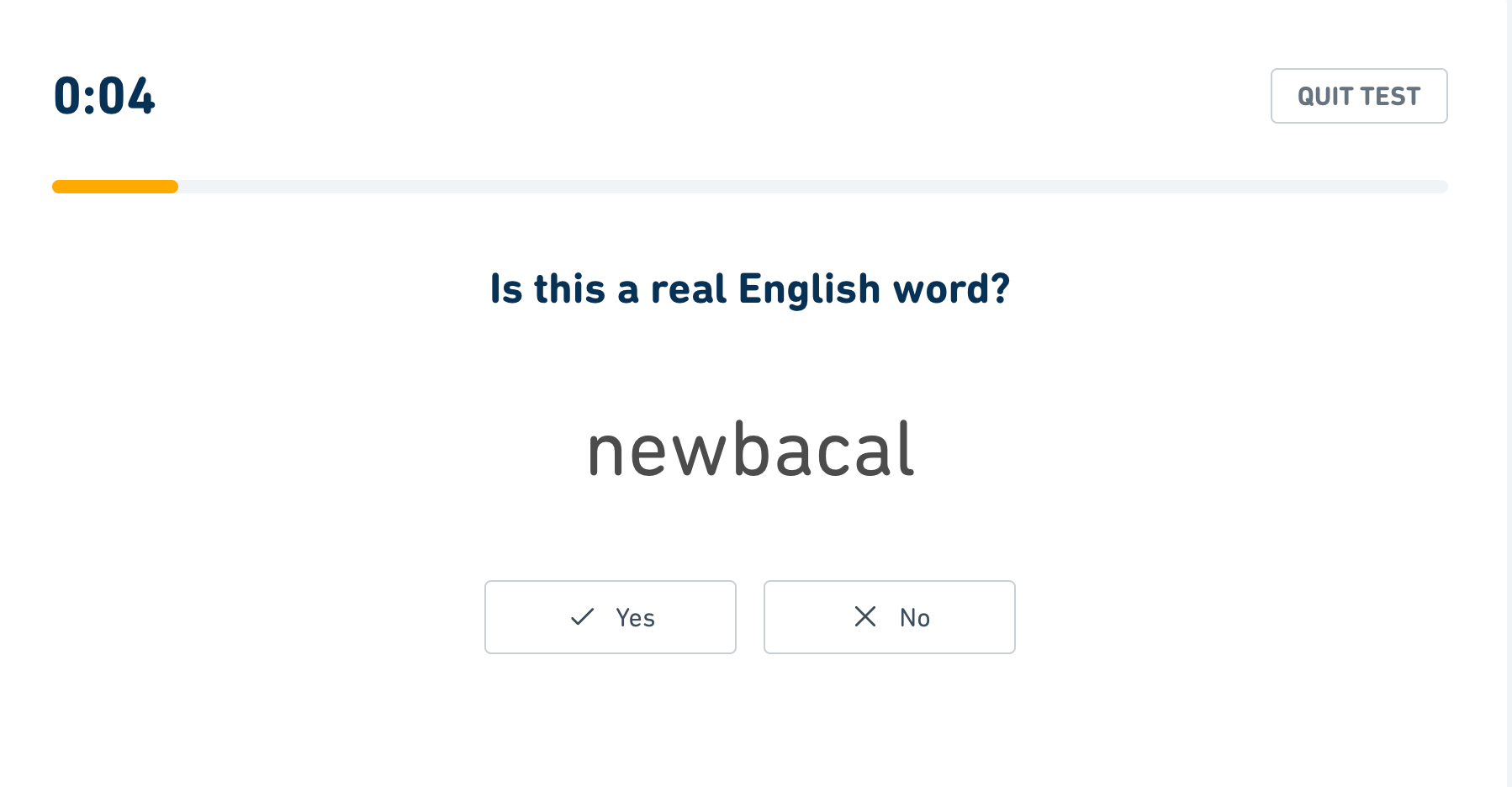}
    \includegraphics[width=0.49\linewidth]{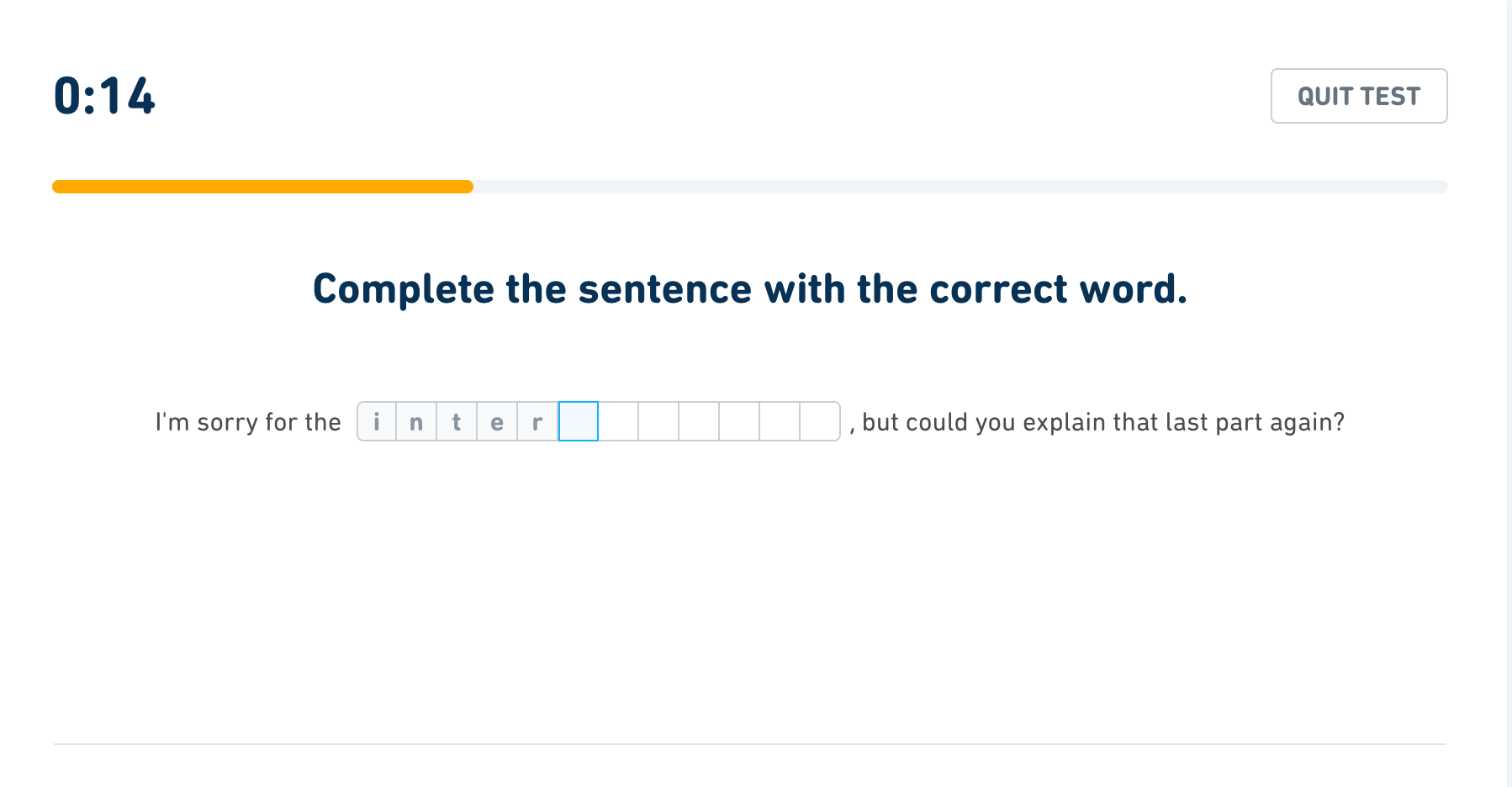}
    \caption{Examples of the yes/no vocabulary and vocab-in-context item types.  These examples are from the DET practice test as of 2024-08-09.}
    \label{fig:item_examples}
\end{figure*}

\begin{figure*}
    \centering
    \includegraphics[width=0.49\linewidth]{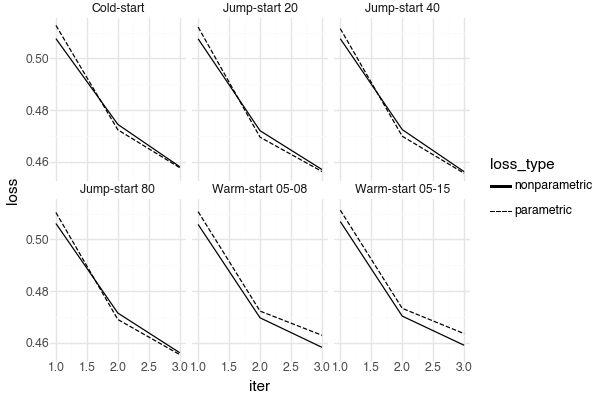}
    \includegraphics[width=0.49\linewidth]{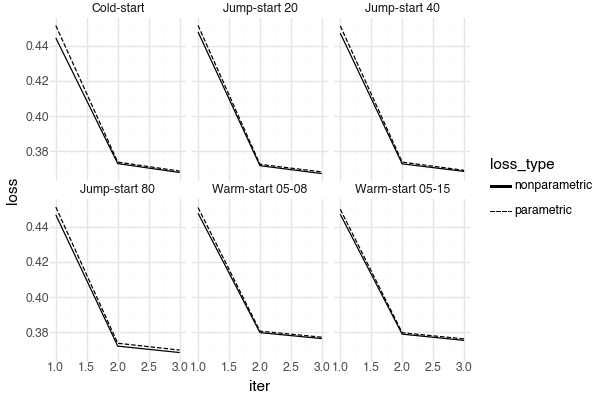}
    \caption{AutoIRT training losses per EM iterations for Y/N Vocab items (left) and ViC items (right).  We see that for most of these experiments the final AutoIRT M-step predictions (parametric) has comparable loss to the intermediate AutoML predictions (non-parametric).}
\end{figure*}

\end{document}